\pdfoutput=1

\documentclass[11pt]{article}

\usepackage[]{EMNLP2022}

\usepackage{times}
\usepackage{latexsym}

\usepackage[T1]{fontenc}

\usepackage[utf8]{inputenc}

\usepackage{microtype}

\usepackage{inconsolata}

\usepackage{epstopdf}
\usepackage[utf8]{inputenc}
\usepackage{tabularx,colortbl}
\usepackage{multirow}
\usepackage{arydshln}
\usepackage{booktabs}
\usepackage{subfigure}
\usepackage{hyperref}
\usepackage{xstring}
\usepackage{amsthm,amsmath,amssymb}
 \usepackage{algorithm}
\usepackage{algpseudocode}
\usepackage{graphicx}
\usepackage[normalem]{ulem}
\usepackage{pinyin}
\usepackage{mathrsfs}
\usepackage{CJK}
\usepackage{enumitem}
\PassOptionsToPackage{hyphens}{url}\usepackage{hyperref}

\title{DuReader$_{\bf retrieval}$: A Large-scale Chinese Benchmark for Passage Retrieval from Web Search Engine}

\author{Yifu Qiu$^1$\footnotemark[2], Hongyu Li$^2$, Yingqi Qu$^2$, Ying Chen$^2$, \textbf{Qiaoqiao She}$^2$, \\ \textbf{Jing Liu}$^2$*, \textbf{Hua Wu}$^2$, \textbf{Haifeng Wang}$^2$ \\
        $^1$Institute for Language, Cognition and Computation, University of Edinburgh, UK \\
        $^2$Baidu Inc., Beijing, China \\
        y.qiu-20@sms.ed.ac.uk \\
         \{lihongyu04, quyingqi, chenying04, sheqiaoqiao,  liujing46, wu\_hua, wanghaifeng\}@baidu.com \\
        }

\begin{document}
\maketitle

\renewcommand{\thefootnote}{\fnsymbol{footnote}}
\footnotetext[2]{The work was done when the first author was doing internship at Baidu.} 
\footnotetext[1]{Corresponding author.}

\begin{abstract}
In this paper, we present DuReader$_{\bf retrieval}$, a large-scale Chinese dataset for passage retrieval. DuReader$_{\bf retrieval}$ contains more than 90K queries and over 8M unique passages from a commercial search engine. To alleviate the shortcomings of other datasets and ensure the quality of our benchmark, we (1) reduce the false negatives in development and test sets by manually annotating results pooled from multiple retrievers,  and (2) remove the training queries that are semantically similar to the development and testing queries.
Additionally, we provide two out-of-domain testing sets for cross-domain evaluation, as well as a set of human translated queries for for cross-lingual retrieval evaluation. 
The experiments demonstrate that DuReader$_{\bf retrieval}$ is challenging and a number of problems remain unsolved, such as the salient phrase mismatch and the syntactic mismatch between queries and paragraphs. 
These experiments also show that dense retrievers do not generalize well across domains, and cross-lingual retrieval is essentially challenging. 
DuReader$_{\bf retrieval}$ is publicly available at 
\url{https://github.com/baidu/DuReader/tree/master/DuReader-Retrieval}.
\end{abstract}

\section{Introduction}

Passage retrieval requires systems to select candidate passages from a large passage collection. 
In recent years, pre-trained language models~\cite{devlin-etal-2019-bert,DBLP:journals/corr/abs-1907-11692RoBERTa} have been applied to retrieval problems, known as \textit{dense retrieval}~\cite{karpukhin-etal-2020-dense-passage-retrieval,qu-etal-2021-rocketqa,10.1145/3404835.3462880-optimizing-dense-retriever-with-hard-negatives}. 
The success of dense retrieval relies on the availability of high quality, large-scale, human-annotated corpora. 
A number of popular datasets are already available for English passage retrieval, including MS-MARCO \citep{nguyen2016msMARCO}, TriviaQA \cite{joshi-etal-2017-triviaqa}, and Natural Questions \citep{kwiatkowski-etal-2019-natural-question}. 
In contrast, existing datasets for non-English retrieval (e.g., Chinese), are either small or machine generated.
For example, TianGong-PDR \citep{wu2019investigating} has only 70 questions and 11K passages. 
Even though the multilingual dataset mMARCO\citep{bonifacio2021mmarco} is large in size, it is constructed by machine translation from the English MS-MARCO dataset.
Sougou-QCL \cite{10.1145/3209978.3210092-sougou-qcl-clickingmodel} is constructed based on click logs of web data without human annotation. 
In this paper, we present DuReader$_{\bf retrieval}$, a large-scale Chinese dataset for passage retrieval from web search engine, that is manually annotated. 
The dataset contains more than 90K queries and over 8M unique passages. All queries are selected from real requests made by users at Baidu Search, and document passages are from the search results. Similar to \cite{karpukhin-etal-2020-dense-passage-retrieval}, we create the DuReader$_{\bf retrieval}$ from DuReader \cite{he-etal-2018-dureader}, a Chinese machine reading comprehension dataset, and obtain the human labels for paragraphs by distant supervision (See Section \ref{sec:dataset-construction}). 
An example from DuReader$_{\bf retrieval}$ is shown in Table~\ref{tab:case_study}, and a comparison of different datasets is shown in Table~\ref{tab:compare-other-data}. 

\begin{table*}[t]
\scalebox{0.49}{
\begin{tabular}{p{32cm}}
\toprule
\textbf{Query}:\\ \begin{CJK}{UTF8}{gbsn} 太阳花怎么养\end{CJK} \\ How to raise Grandiflora? \\ \midrule
\textbf{Positive Psg. 1}        :\\   \begin{CJK}{UTF8}{gbsn}百度经验:jingyan.baidu.com花卉名称:太阳花播种时间:春、夏、秋均可播种为一年生肉质草本植物。株高10～15cm。花瓣颜色鲜艳,有白、深黄、红、紫等色。园艺品种很多,有单瓣、半重瓣、重瓣之分。喜温暖、阳光充足而干燥的环境,极耐瘠薄,一般土壤均能适应,能自播繁衍。见阳光花开,早、晚、阴天闭合,故有太阳花、午时花之名。花期6～7月。太阳花种子非常细小。常采用育苗盘播种,极轻微地覆些细粒蛭石,或仅在播种后略压实,以保证足够的湿润。发芽温度21～24℃,约7～10天出苗,幼苗极其细弱,因此如保持较高的温度,小苗生长很快,便能形成较为粗壮、肉质的枝叶。这时小苗可以直接上盆,采用10厘米左右直径的盆,每盆种植2～5株,成活率高,生长迅速。\end{CJK}     \\ Baidu experience: jingyan.baidu.com Flower name: Grandiflora Sowing time: Spring, summer, and autumn can be sown as an annual succulent herb. Plant height is 10-15cm. The petals are bright in color, white, dark yellow, red, purple and other colors. There are many horticultural varieties, including single, semi-double and double petals. It likes a warm, sunny and dry environment, is extremely tolerant to barrenness, and can adapt to general soils and reproduce by itself. It is named Grandiflora because it blooms when the sun is rising and closes in the morning, evening and cloudy days. It flowers from June to July. 
Grandiflora's seeds are very small. The seedling trays are often used for sowing, very lightly covered with fine vermiculite, or only slightly compacted after sowing to ensure sufficient moisture. The germination temperature is 21\textasciitilde24$^{\circ}$C and the seedlings emerge in about 7\textasciitilde10 days. The seedlings are extremely thin. Therefore, if the temperature is kept high, the seedlings will grow quickly, and thicker, fleshy branches and leaves can be formed. At this time, the seedlings can be directly put into pots, using pots with a diameter of about 10 cm, planting 2 to 5 plants per pot, with high survival rate and rapid growth.               \\ \midrule
\textbf{Positive Psg. 2}        :\\   \begin{CJK}{UTF8}{gbsn}抹平容器中培养土平面,将剪来的太阳花嫩枝头插入竹筷戳成的洞中,深入培养土最多不超过2厘米。为使盆花尽快成形、丰满,一盆中可视花盆大小,只要能保持2厘米的间距,可扦插多株(到成苗拥挤时,可分栽他盆)。接着浇足水即可。新扦插苗可遮阴,也可不遮阴,只要保持一定湿度,一般10天至15天即可成活,进入正常的养护。太阳花极少病虫害。平时保持一定湿度,半月施一次千分之一的磷酸二氢钾,就能达到花大色艳、花开不断的目的。如果一盆中扦插多个品种,各色花齐开一盆,欣赏价值更高。每年霜降节气后(上海地区)将重瓣的太阳花移至室内照到阳光处。入冬后放在玻璃窗内侧,让盆土偏干一点,就能安全越冬。次年清明后,可将花盆置于窗外,如遇寒流来袭,还需入窗内养护。\end{CJK} \\
Flatten the soil surface in the container, insert the cut branches of Grandiflora into the hole made by the bamboo chopsticks, and deepen the soil for no more than 2 cm. To make the potted flowers take shape and fullness as soon as possible, multiple plants can be cut as long as the spacing of 2 cm can be maintained (when the seedlings are crowded, they can be planted in other pots). Then pour plenty of water. The new cuttings can be shaded or not. As long as they maintain a certain humidity, they can survive 10 to 15 days and enter normal maintenance. Grandiflora has very few pests and diseases. Maintain a certain humidity at ordinary times, and apply one-thousandth of potassium dihydrogen phosphate once a half month to achieve the purpose of large flowers and continuous blooming. If there are multiple varieties of cuttings in one pot, the flowers of all colours will bloom in one pot, and the appreciation value will be higher. Every year after the frost falls (Shanghai area), the double-flowered Grandiflora is moved indoors to shine in the sun. Put it on the inside of the glass window after the winter, and let the potting soil dry a little to survive the winter safely. After the Qingming Festival in the following year, the flowerpots can be placed outside the window.
    \\ \bottomrule
\end{tabular}%
}
\caption{A data instance randomly selected from the DuReader$_{\bf retrieval}$ development set.}
\label{tab:case_study}
\end{table*}

Additionally, recent works point out two major shortcomings of the development and testing sets in the existing datasets: 
\begin{itemize}[noitemsep,leftmargin=*]
    \item \citet{arabzadeh2021shallowPoolingSparseLabel} and \citet{qu-etal-2021-rocketqa} observe that false negatives (i.e. relevant passages but labeled as negatives) are common in the passage retrieval datasets due to their large scale but limited human annotation. As a result, the top passages retrieved by models may be superior to labeled relevant positives, and this will affect the evaluation. 
    \item \citet{lewis-etal-2021-TestTrainOverlapODQA} find that 30\% of the test-set queries in the common machine reading comprehension datasets \cite{kwiatkowski-etal-2019-natural-question,joshi-etal-2017-triviaqa} have a near-duplicate paraphrase in their corresponding training sets, thus leaking the testing information into model's training. The similar issue has been observed in MSMARCO~\cite{Zhan2022EvaluatingEP}. 
\end{itemize}
In the construction of DuReader$_{\bf retrieval}$, we try to alleviate the above issues and improve the quality of the development and testing sets in the following two ways (see Section~\ref{sec:quality-improvement}): 
\begin{itemize}[noitemsep,leftmargin=*]
 
\item To reduce the false negatives in the development and testing set, we invite the internal data team to manually check and relabel the passages in the top retrieved results pooled from multiple retrievers. 
\item
To reduce the leakage of testing information into model's training, we use a query matching model from \cite{zhu2021duqm} to identify and remove the training queries that are semantically similar to the development and testing queries.

\end{itemize}
Moreover, inspired by \citet{thakur2021beir}, we provide two testing sets (see Section~\ref{sec:out-of-domain-set}) from the medical domain (cMedQA \cite{cmedq} \footnote{avaliable at \href{https://github.com/zhangsheng93/cMedQA2}{https://github.com/zhangsheng93/cMedQA2}} and cCOVID-News\footnote{available at \href{https://www.datafountain.cn/competitions/424}{https://www.datafountain.cn/competitions/424}}) as the separate testing sets for out-of-domain evaluation. Additionally, we also provide a set that contains human translated queries for cross-lingual retrieval evaluation (see Section~\ref{sec:cross-lingual-set}) \cite{asai-etal-2021-xor-tidy,sun-duh-2020-clirmatrix}. 

In this paper, we conduct extensive experiments. 
In our in-domain experiments, we find that there are many challenges to be addressed,  such as salient phrase mismatches and syntactic mismatches (see Section~\ref{sec:in-domain-challenges}). It is also difficult for dense retrievers to generalize well across different domains as we observed in the out-of-domain experiments (see Section~\ref{sec:out-of-domain-experiment}). Finally, the cross-lingual experiments indicate that cross-lingual retrieval is essentially challenging (see Section~\ref{sec:cross-lingual-experiment}).

We summarize the characteristics of our dataset and our contributions as follows:
\begin{itemize}
    \item We present a large-scale Chinese dataset for benchmarking the passage retrieval models. Our dataset comprises more than 90K queries and more than 8M unique passages from Baidu Search. 
    \item We leverage two strategies to improve the quality of our benchmark and alleviate the existing shortcomings in other existing datasets, including reducing the false negatives with human annotations on pooled retrieval results, and remove the training queries semantically similar to the development and testing queries.
    \item We introduce two extra out-of-domain test sets to evaluate the domain generalization capability, and a cross-lingual set to assess the cross-lingual retrievers.
    \item We conduct extensive experiments and the results demonstrate that the dataset is challenging and passage retrieval has plenty of room for improvement. 
\end{itemize}

\begin{table*}[]
\centering
\scalebox{0.73}{
\begin{tabular}{ccccccc}
\toprule
Dataset                     & Lang        & \#Que.       & \#Psg.        & Source of Que.     & Source of Psg.  & Psg. Annotation   \\ \midrule
MS-MARCO \cite{nguyen2016msMARCO}                    & EN          & 516K         & 8.8M          & User logs          & Web doc.       & Human    \\
TriviaQA \cite{joshi-etal-2017-triviaqa}                     & EN          & 95K          & 650K          & Trivia web.        & Wiki./Web doc.  &  Dist. Sup.  \\
Natural Questions \cite{kwiatkowski-etal-2019-natural-question}            & EN          & 61K          & 21M           & User logs          & Wiki doc.    &   Dist. Sup.    \\ \midrule
mMARCO-Chinese \cite{bonifacio2021mmarco}              & CN          & 516K         & 8.8M          & User logs      & Web doc.   & Tranlation    \\
cCOVID-News\footnote{available at \href{ www.datafountain.cn/competitions/424}{www.datafountain.cn/competitions/424}}                & CN          & 4.9K         & 5K            & User question     & COVID-19 News doc. & Human \\
cMedQA-2.0 \cite{cmedq}                      & CN          & 108K         & 203K           & User question     & Medical Forum    & Question-Answer Pairs \\
TianGong-PDR \cite{wu2019investigating}                & CN          & 70           & 11K          & User logs          & News doc.       &  Human \\ 
Sougou-QCL \cite{10.1145/3209978.3210092-sougou-qcl-clickingmodel}                      & CN          & 537K         & 9M           & Click logs     & Web data    & Click signal \\ \midrule
\textbf{DuReader$_{\bf retrieval}$} (Our work) & \textbf{CN} & \textbf{97K} & \textbf{8.9M} & \textbf{User logs} & \textbf{Web doc.}  & \textbf{Dist. Sup. + Human} \\ \bottomrule
\end{tabular}}
\caption{Summary of data statistics for passage retrieval datasets. The  annotation of passages in TriviaQA and Natural Questions are presented in \cite{karpukhin-etal-2020-dense-passage-retrieval}. Compared with other works, the instances in DuReader$_{\bf retrieval}$ come from user logs in web search. Its consists of a distant supervised (Dist. Sup.)  training set and human-annotated (Human) development and test sets.} 
\label{tab:compare-other-data}
\end{table*}

\section{DuReader$_{\bf retrieval}$}
In this section, we introduce our DuReader$_{\bf retrieval}$ dataset (See dataset statistics in Table \ref{tab:dataset-statistics}). We first formally define the passage retrieval task in Section \ref{sec:task-define}. We then introduce how we initially construct our dataset from DuReader by distant supervision in Section \ref{sec:dataset-construction}. Our strategies for further improving the data quality are discussed in Section \ref{sec:quality-improvement}. Finally, we introduce two out-of-domain test sets in Section \ref{sec:out-of-domain-set} and a cross-lingual set in in Section \ref{sec:cross-lingual-set}.

\subsection{Task Definition}\label{sec:task-define}
DuReader$_{\bf retrieval}$ is created for the task of passage retrieval, that is, retrieving a list of relevant passages in response to a query. Formally, given a query $q$ and a large passage collection $\mathcal{P}$, a retrieval system $\mathcal{F}$ is required to return the top-$K$ relevant passages  $P_K^{(q)}=\left\{p_1^{(q)},p_2^{(q)},...,p_K^{(q)}\right\}$, where $K$ is a manually defined number.
Ideally, all the relevant passages to $q$ within $\mathcal{P}$ should be included and ranked as high as possible in the retrieved results $P_K^{(q)}$.

\begin{table}[h]
\centering
\scalebox{0.67}{
\begin{tabular}{lrrr} 
\toprule
\textbf{DuReader$_{\bf retrieval}$ }                               & \textbf{Train} & \textbf{Dev.} & \textbf{Test}  \\ 
\midrule
\#Chinese queries                                & 86,395    & 2,000 & 4,000              \\
\#passages                              &  222,395 & 9,863 & 19,601                   \\
\#avg.
  passages per query              &     2.57 & 4.93 & 4.90                \\
\#avg.
  Chinese characters per query  &     9.51 & 9.29 & 9.23                \\
\#avg.
  Chinese characters per passage &    358.58 & 398.59 & 401.61                 \\
  
  \midrule
  \textbf{cMedQA  (Out-of-domain)}              & \textbf{Train} & \textbf{Dev.} & \textbf{Test}  \\ 
\midrule
\#queries                                &   \textbackslash{}  & \textbackslash{} &     3,999         \\
\#passages                              &  \textbackslash{} & \textbackslash{} &    7,527                \\
\#avg.
  passages per query              &     \textbackslash{} & \textbackslash{} &    1.88             \\
\#avg.
  Chinese characters per query  &    \textbackslash{} & \textbackslash{} &    48.47             \\
\#avg.
  Chinese characters per passage &    \textbackslash{} & \textbackslash{}  & 100.58                 \\
 \midrule
   \textbf{cCOVID-News (Out-of-domain)}                                & \textbf{Train} & \textbf{Dev.} & \textbf{Test}  \\ 
\midrule
\#queries                                & \textbackslash{}    & \textbackslash{} & 949              \\
\#passages                              &  \textbackslash{} & \textbackslash{} &    964                \\
\#avg.
  passages per query              &     \textbackslash{} & \textbackslash{} &    1.02             \\
\#avg.
  Chinese characters per query  &     \textbackslash{} &  \textbackslash{}  & 25.93                \\
\#avg.
  Chinese characters per passage &    \textbackslash{} & \textbackslash{}  & 1430.93                 \\
\midrule
  \multicolumn{3}{l}{\textbf{Translated queries for cross-lingual retrieval}} \\
  \midrule
  \#avg.
  English words per query  &     6.41 & 6.55 & 6.46                \\
  \#English queries                                & 9,500    & 2,000 & 4,000              \\
\midrule
Size of the total paragraph collection &     & \centering 8,096,668  &  \\
\bottomrule
\end{tabular}}
\caption{Summary of statistics for the training (Train), development (Dev.), testing (Test), out-of-domain (OOD) testing sets and cross-lingual set of DuReader$_{\bf retrieval}$.}
\label{tab:dataset-statistics}
\end{table}

\subsection{Dataset Construction}
\label{sec:dataset-construction}

\subsubsection{An Introduction of DuReader Dataset}
DuReader$_{\bf retrieval}$ is developed based on the Chinese machine reading comprehension dataset DuReader \cite{he-etal-2018-dureader}. All queries in DuReader are posed by the users of our chosen commercial search engine, and document-level contexts are gathered from search results. Each instance in DuReader is a tuple $<q,t,D,A>$, where $q$ is a query, $t$ is a query type, $D$ is the top-5 retrieved documents constituted by their paragraphs returned by our chosen commercial search engine. $A$ is the answers written by human annotators.

\subsubsection{Constructing DuReader$_{\bf retrieval}$ from DuReader} 

In this section, we describe that how we construct DuReader$_{\bf retrieval}$ from DuReader. First, we describe our approach to labelling the positive passages. Then, we discuss our approaches to dealing with the two challenges in constructing DuReader$_{\bf retrieval}$ from DuReader: 1) the original paragraphs are too short to provide meaningful context; and 2) the term overlap between the queries and the document titles may ease the challenges for passage retrieval. 

\noindent \textbf{Distant Supervision for Annotations} \quad Following MS-MARCO Passage Ranking \cite{nguyen2016msMARCO}, we use the human-written answers from DuReader \cite{he-etal-2018-dureader} to label the positive passages by the distant supervision. A paragraph is considered positive if it contains any human-written answer. 
Specifically, we leverage the span-level F1 score to measure the match between each human-written answer and the paragraphs in documents. 
If a span-answer pair gets a F1-score higher than the threshold (0.5), we label the paragraph as positive. 
We show the details of our annotation process in Algorithm \ref{alg:find-pos-para}.

\begin{algorithm}
\caption{Span-level F1 Annotation for Positives}\label{alg:find-pos-para}
\textbf{Input:} $\{ \langle p,a \rangle \}$, $p$: candidate paragraph, $a$: answer, $\tau$: threshold for positive labelling.\\
\textbf{Output:} $l_p\in\{0,1\}$: label, $0$ and $1$ denote negative and positive $p$ , separately. 
\begin{algorithmic}

\For{any span $s$ in $p$}{
\If{Calculate $F1(s,a) \geq \tau$}:
    \State $l_p \gets 1$
    \State \textbf{return}
\EndIf
}\EndFor
\State $l_p \gets 0$ \\
\textbf{return}
\end{algorithmic}
\end{algorithm}

\noindent \textbf{Passage Length Control} \quad
Additionally, most paragraphs in DuReader are too short to form meaningful contexts. We concatenate the paragraphs of each document in DuReader by the following rules: 1) In a document of less than 256 Chinese characters, all paragraphs are concatenated into one passage; 2) In a document of more than 256 Chinese characters, a paragraph of less than 256 is concatenated with the next one, and the concatenation does not stop until the length of the new passage exceeds 256. The new passage is labelled as positive if any of its components are originally labelled positive in DuReader. After the processing, the median and the mean of the passage length are 304 and 272, respectively. 

\noindent \textbf{Removing Document Titles} \quad
We remove the titles from all documents in DuReader, since we observe that there is many term overlaps between the queries and the titles. 
If we keep them, the retrieval systems may easily match the queries with the document titles and achieve high performance. 
But we expect the retrievers to capture all contextual information in passages to answer queries. 

\subsection{Quality Improvement}
\label{sec:quality-improvement}

As we discussed in the previous section, there are shortcomings of other existing datasets. 
To alleviate such shortcomings, we further design two strategies to ensure the quality of the development and test sets in DuReader$_{\bf retrieval}$. Although in this work we apply our quality improvement approaches to the Chinese passage retrieval dataset, the proposed method allows flexibility extended to other languages (e.g., English), benefiting the future evaluation and development of dense retrieval systems.

\noindent \textbf{Reducing False Negatives} \quad 
A common issue in existing passage retrieval datasets~\cite{qu-etal-2021-rocketqa, arabzadeh2021shallowPoolingSparseLabel} is false negatives, i.e., query-relevant passages not labelled as positives, in the development and testing sets. 
In this section, we discuss our strategy for reducing the false negatives in the development and testing sets of DuReader$_{\bf retrieval}$.

We use human annotation as a complement to the distant supervised labeling approach discussed in Section \ref{sec:dataset-construction}. We invite the internal data team to manually check the labels in the development and test sets and fix them if necessary. 
To avoid inductive bias in our annotation process, we follow the pooling method in TREC competitions~\cite{voorhees2005trec} to select candidate passages for annotation. 
The top-ranked passages retrieved for each query by a set of \textit{contributing retrievers} are pooled for annotation. 
In particular, the annotator is presented with a query and the top-5 passages pooled from five retrieval systems. 
We use BM25 and four neural retrievers with the initialization from ERNIE \cite{sun2019ernie}, BERT \cite{devlin-etal-2019-bert, cui-etal-2021-pretrain}, RoBERTa \cite{DBLP:journals/corr/abs-1907-11692RoBERTa} and MacBERT \cite{cui-etal-2020-revisiting-MacBERT}  to serve as our contributing retrievers. 
We combine their top-50 retrieved passages as candidates. 
An ensembled re-ranker is then used (See Appendix \ref{appendix:pooling-details} for implementation details) to select the top-5 passages for human annotation. 
To ensure data quality, we perform all annotations on our internal annotation platform. 
Please refer to the Appendix~\ref{appendix:crosslingual-set-details} for annotation settings and quality control.

After adopting our strategy for reducing false negatives, the average positive paragraph per query has increased from $2.43$ to $4.91$. 
$71.53\%$ of queries have at least one false negative relabeled by annotators, 
which shows there are many false negatives in the raw corpus derived directly from DuReader.

\noindent \textbf{Removing Similar Queries} \quad Retrieval systems should avoid merely memorizing queries and their relevant items in the training set and directly applying such memorization during inference.  \citet{lewis-etal-2021-TestTrainOverlapODQA} find that in some popular datasets, including Natural Questions \cite{kwiatkowski-etal-2019-natural-question}, WebQuestions \cite{berant-etal-2013-webquestions} and TriviaQA \cite{joshi-etal-2017-triviaqa}, 30\% of the test-set queries have a near-duplicate paraphrase in their corresponding training sets, which leaks the testing information into the model training. 
In this paper, we use a model-based approach to remove training queries that are semantically similar to development and testing queries.

We use the query matching model in \cite{zhu2021duqm}, which computes the similarity score ranging between $[0,1]$ for a query pair.
We set a threshold of 0.5, meaning that if the similarity between a training query and a test query is higher than 0.5, we mark the query pair as semantically similar. 
There are 566 training queries semantically similar to 387 queries in the development and the test set, accounting for approximately 6.45\% of total development and test queries. 
All these 566 training instances are removed in DuReader$_{\bf retrieval}$. 

\subsection{Out-of-domain Evaluation}\label{sec:out-of-domain-set}
Recent work~\cite{thakur2021beir} reveals that the dense retrievers do not generalize well cross-domain. 
To assessing the cross-domain generalization ability of the retrievers, we carefully select two publicly available Chinese text retrieval datasets, i.e., cMedQA \cite{cmedq} created from online medical consultation text and cCOVID-News from COVID-19 news articles. 
We randomly select 949 and 3,999 samples from cCOVID-News and cMedQA, respectively, as out-of-domain testing data. 

\subsection{Cross-lingual Evaluation} \label{sec:cross-lingual-set}
Cross-lingual passage retrieval has recently received much attention \cite{shi-etal-2021-crosslingual-dense-document-retrieval,NEURIPS2021_3df07fda-one-question-answering-model-from-many-language-crosslingual}, which aims to retrieve the passages in the target language (e.g., Chinese) as the response to the query in source language (e.g., English).

In DuReader$_{\bf retrieval}$, we provide a cross-lingual retrieval set which contains the English queries paired with Chinese positive passages. 
The total numbers of training/development/testing English queries are 9.5K/4K/2K, respectively.
All English queries in our cross-lingual set are translated and the passage annotations are aligned with DuReader$_{\bf retrieval}$. 
To obtain English queries, we first translate Chinese queries to English queries by using machine translation~\footnote{\url{https://fanyi.baidu.com}}. 
Then, we ask the internal data team to manually check the quality of the machine-translated queries, and modify the translated queries if necessary. The quality controls for translated queries are the same as our previous human annotations for the in-domain development and testing set as in Appendix \ref{appendix:crosslingual-set-details}.

\begin{figure}
    \centering
    \includegraphics[width=\linewidth]{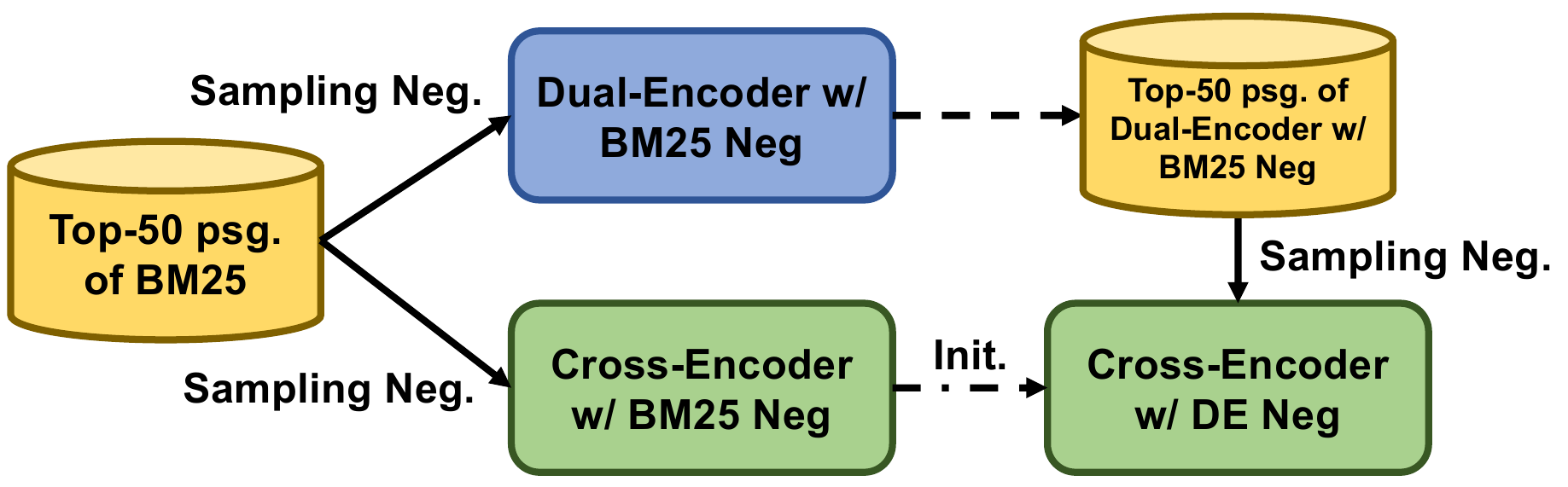}
    \caption{Illustration for the training procedure of our one dual-encoder retriever and two cross-encoder re-rankers. We train our first retriever and re-ranker by the negatives sampled from BM25's output as in \cite{karpukhin-etal-2020-dense-passage-retrieval}. We further attempt the strategy in \cite{DBLP:conf/iclr/XiongXLTLBAO21-ANCE} that sampling negatives from dual-encoder retriever to enhance the cross-encoder re-ranker.}
    \label{fig:baseline}
\end{figure}

\section{Experiments and Results}
\label{sec:experiments-results}
\subsection{Baselines}\label{sec:baselines}

We use the recent two-stage framework (retrieve-then-rerank) \cite{10.1007/978-3-642-36973-5_36-two-stage-learning-to-rank,qu-etal-2021-rocketqa} for passage retrieval and evaluate two retrieval and two reranking models on our DuReader$_{\bf retrieval}$ dataset. 
In particular, we utilize the dual-encoder and cross-encoder architecture in RocketQA~\cite{qu-etal-2021-rocketqa} to develop our neural retrievers and re-rankers. We introduce the baselines as follows. 

\noindent \textbf{BM25} BM25 is a sparse retrieval baseline~\cite{robertson2009probabilisticRelevanceFramework-bm25}. 

\noindent \textbf{DE w/ BM25 Neg} \quad 
\citet{karpukhin-etal-2020-dense-passage-retrieval} shows that the hard negatives from BM25 are more effective at training the dense retrievers than in-batch random negatives. 
With BM25's hard negatives, we train a dual-encoder as our first neural retriever. 

\noindent \textbf{CE w/ BM25 Neg} \quad We use BM25's hard negatives to train a cross-encoder as our first neural re-ranker. 

\noindent \textbf{CE w/ DE Neg} \quad 
CE w/ DE Neg is the second enhanced re-ranker. 
We follow \citet{qu-etal-2021-rocketqa} to train CE w/ DE Neg. Specifically, we use CE w/ BM25 Neg to initialize the parameters, and use DE w/ BM25 Neg to retrieve negatives from the entire passage collection. 

The relationships among our neural retrievers and re-rankers are shown in Figure \ref{fig:baseline}. The training and architectural settings for all models are detailed in the Appendix \ref{appendix:implmentation-details}.

\subsection{Evaluation Metrics}

We use the following evaluation metrics in our experiments: (1) Mean Reciprocal Rank for the top 10 retrieved documents (MRR@10), (2) Recall for the top-1 retrieved items (Recall@1) and (3) Recall for the top-50 retrieved items (Recall@50). Recall@50 is more suitable for evaluating the first-stage retrievers, while MRR@10 and Recall@1 are more suitable for assessing the second-stage re-rankers. 

\subsection{Baseline Performance}

\begin{table}[]
\centering
\scalebox{0.77}{
\begin{tabular}{lccc}
\toprule
                  & \textbf{MRR@10} & \textbf{Recall@1} & \textbf{Recall@50} \\ \midrule
BM25              & 21.03           & 12.08             & 70.00               \\
DE w/ BM25 Neg & \textbf{53.96}           & \textbf{41.53}	            & \textbf{91.33}              \\
\bottomrule
\end{tabular}}
\caption{Performance of retrieval models on the testing set of DuReader$_{\bf retrieval}$.}
\label{tab:main-eval-retrieval}
\end{table}

\begin{table}[]
\centering
\scalebox{0.80}{
\begin{tabular}{lccc}
\toprule
\textit{BM25's top-50 psg.}               & \textbf{MRR@10} & \textbf{Recall@1} & \textbf{Recall@50} \\ \midrule
CE w/ BM25 Neg        & 56.80           & 48.83             & 70.00               \\
CE w/ DE Neg          & 57.62           & 51.52              & 70.00              \\ \midrule
\textit{DE's top-50 psg.} & \textbf{MRR@10} & \textbf{Recall@1} & \textbf{Recall@50} \\ \midrule
CE w/ DE Neg          & \textbf{74.21}  & \textbf{66.03}    & \textbf{91.33}     \\ \bottomrule
\end{tabular}}
\caption{Performance of re-ranking models on testing set of DuReader$_{\bf retrieval}$. We present re-ranking results based on two retrieval models including \textit{BM25} and \textit{DE w/ BM25 Neg}.}
\label{tab:main-eval-rerank}
\end{table}

We report the in-domain baseline performances for the first-stage retrievers in Table \ref{tab:main-eval-retrieval}. Compared with the traditional retrieval system BM25, it is expected that DE w/ BM25 Neg outperforms the traditional system among all metrics, thanks to the powerful expressive ability of the neural encoder.

We then report the in-domain baseline performances for the second-stage re-rankers in Table \ref{tab:main-eval-rerank}. We observe that training the re-ranker with the hard negatives sampled from the neural retriever's top predictions is shown to outperform the negatives sampled from BM25's retrieved results in terms of MRR@10 and Recall@1. 

\subsection{Effects of Quality Improvements}

In this section, we examine the effects of our strategies to improve the data quality of DuReader$_{\bf retrieval}$ as in Section \ref{sec:quality-improvement}.

\begin{figure*}
    \centering
    \includegraphics[width=0.9\linewidth]{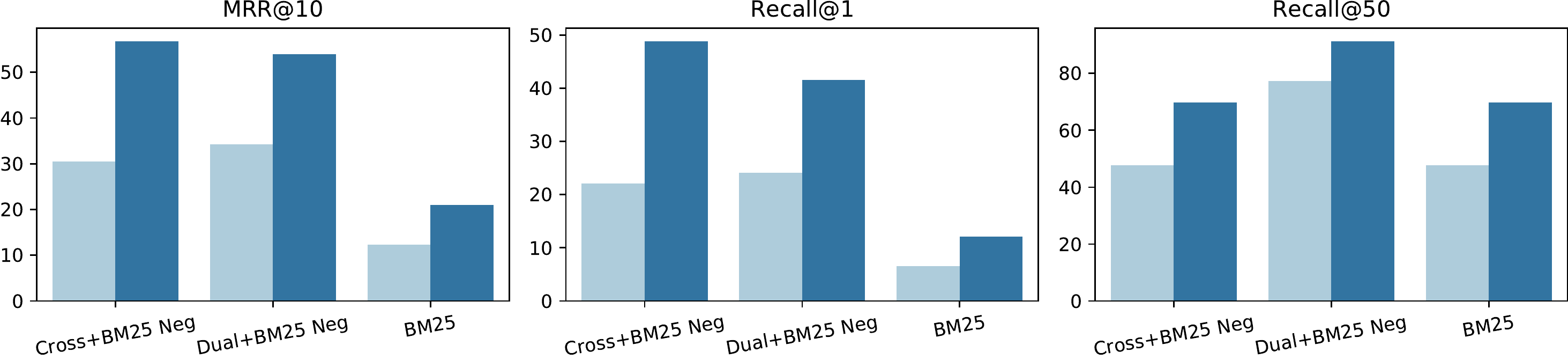}
    \caption{Comparison in the model performances before (light blue) and after (dark blue) reducing the false negatives in the development set via our pooling strategy and human annotation.}
    \label{fig:reducing-false-negative}
\end{figure*}

\noindent \textbf{Reducing False Negatives} \quad We test three models, including BM25, a dense retrieval model (DE w/ BM25 Neg) and a re-ranking model (CE w/ BM25 Neg) based on BM25's top-50 retrieved results, to quantify the impact of our strategy on reducing false negatives. 
Specifically, we compare the performance of the same model on the development set either with or without reducing false negatives. 
As shown in Figure \ref{fig:reducing-false-negative}, all metrics of the three models are significantly improved after adopting our strategy. 
These results suggest that there are many false negatives in the raw retrieval dataset derived from DuReader, and that our strategy successfully captures and reduces false negatives in development and testing sets.

\begin{table}[]
\centering
\scalebox{0.8}{
\begin{tabular}{ccc}
\toprule
\multirow{2}{*}{\textbf{Model}} & \multicolumn{1}{c}{\textbf{MRR@10}}         & \multicolumn{1}{c}{\textbf{Recall@1}}            \\ \cline{2-3} 
                                & \textbf{Duplicated} &  \textbf{Duplicated}   \\ \midrule
CE w/ Sim. Q                 & \textbf{50.6}      &  \textbf{43.93}                                    \\
CE w/o Sim. Q                  & 49.94              &  42.89            \\ \bottomrule
\end{tabular}}
\caption{Comparison of models by using two groups of training data: 1) \textbf{CE w/ Sim. Q}: training data without removing the queries that are semantically similar to the development and testing queries, 2) \textbf{CE w/o Sim. Q}: training data with removing the queries that are semantically similar to the development and testing queries. We evaluate the two models on the duplicated queries (\textbf{Duplicated}). All top-50 retrieval results are based on BM25. We \textbf{bold} the best model on each column.}
\label{tab:qq-overlap}
\end{table}

\noindent \textbf{Removing Similar Queries} We conduct an experiment to quantify the effects of removing the training queries that are semantically similar to the development and testing queries. 
We train our re-ranking model (CE w/ BM25 Neg) by using the training data without (\textbf{CE w/o Sim. Q}) and with (\textbf{CE w/ sim. Q}) semantically duplicated queries, respectively. 
We then test both models on all 387 semantically duplicated queries (\textbf{Duplicated}) in the development and testing sets, as well as the rest of the development set (\textbf{Others}). 
We use BM25's top-50 retrieved results for the re-ranking models to re-rank.
As shown in Table \ref{tab:qq-overlap}, comparing the two models' performance on Duplicated, we find model trained with those semantically similar queries (CE w/ Sim. Q) has a higher score on both MRR@10 and Recall@1. 
This suggests that using semantically similar queries in training may allow the model to simply memorize the data during training and achieve better performance during testing.

\subsection{The Challenges and Limitations} \label{sec:in-domain-challenges}

In this section, we analyze the results of our best baseline system (i.e., retrieving the top-50 passages by DE w/ BM25 Neg, then re-ranking by CE w/ DE Neg) to better understand the specific challenges and limitations of DuReader$_{\bf retrieval}$. Specifically, we manually analyze 500  query-passage predictions of the baseline.
The 500 query-passage pairs are from 100 random-selected development queries with the top-5 passages retrieved and re-ranked by the baseline. 
To help understand the challenges and limitations of DuReader$_{\bf retrieval}$, we ensure that the top-5 passages of these 100 queries contain no positive passages.

\noindent \textbf{Salient Phrase Mismatch} \quad We observe that the mismatch in salient phrases between the query and the retrieved passages is particularly challenging for the baseline system as found in \cite{DBLP:journals/corr/abs-2110-06918-salient-phrase-aware-dense-retrieval}, accounting for 53.4\% of total incorrect predictions. We further divide  \textit{salient phrase} into several sub-categories, i.e., \textit{entity}, \textit{numeral}, and \textit{modifier}. Examples and explanations are in Table \ref{tab:error-analysis} in Appendix \ref{sec:case-error-analysis}. 

\noindent \textbf{Syntactic Mismatch} We also observe that around 1\% predictions have a syntactic mismatch between the query and the passage. The case in Table \ref{tab:error-analysis} in Appendix \ref{sec:case-error-analysis} suggests that it is difficult for the baseline system to ensure the consistency in syntactic relationship between the query and the passages.

\noindent \textbf{Other Challenges} \quad We also show two other typical challenges accounting for 22.6\% incorrect predictions: 1) \textit{Over-sensitivity on term overlap}: whether the baseline system is over-sensitive to retrieve the negative passages that contains a few lexical overlap with queries. 
2) \textit{robustness on typo}: whether the baseline system is robust against typos in queries or passages. Note that our dataset is constructed from the real query log of a commercial search engine. The noise in data (e.g. typos) challenges the robustness of the baseline system.

\noindent \textbf{Limitations in False Negatives} \quad We notice that there are still about 14.8\% false negatives. This suggests that despite the success of our strategy in Section \ref{sec:quality-improvement} to reduce false negatives in development and testing sets to some extents, the presence of false negatives remains a challenge in building a high-quality passage retrieval benchmark. 

\begin{table}[]
\centering
\scalebox{0.82}{
\begin{tabular}{lcccccc}
\toprule
\multicolumn{1}{c}{\multirow{2}{*}{\textbf{cMedQA}}} & \multicolumn{2}{c}{\textbf{MRR@10}} & \multicolumn{2}{c}{\textbf{Recall@1}} & \multicolumn{2}{c}{\textbf{Recall@50}} \\ \cline{2-7} 
\multicolumn{1}{c}{}                        & ZS.    & FT.               & ZS.     & FT.                & ZS.      & FT.                \\ \midrule
BM25                                        & 6.26   & \textbackslash{}  & 4.98    & \textbackslash{}   & 14.05    & \textbackslash{}   \\
DE                                & 4.39   & 15.22             & 2.93    & 11.28              & 16.4     & 40.36              \\
 \bottomrule
\end{tabular}}
\caption{Comparison for BM25 and the dense retrieval model DE w/ BM25 Neg (DE) in the Zero-shot (ZS.) experiments, and fine-tuning (FT.) experiments for estimating the upper-bound performance on cMedQA out-of-domain testing set.}
\label{tab:out-of-domain-eval-cmedqa}
\end{table}

\begin{table}[]
\centering
\scalebox{0.72}{
\begin{tabular}{lcccccc}
\toprule
\multicolumn{1}{c}{\multirow{2}{*}{\textbf{cCOVID-News}}} & \multicolumn{2}{c}{\textbf{MRR@10}} & \multicolumn{2}{c}{\textbf{Recall@1}} & \multicolumn{2}{c}{\textbf{Recall@50}} \\ \cline{2-7} 
\multicolumn{1}{c}{}                             & ZS.    & FT.               & ZS.     & FT.                & ZS.      & FT.                \\ \midrule
BM25                                             & 57.49  & \textbackslash{}  & 48.37   & \textbackslash{}   & 85.67    & \textbackslash{}   \\
DE                                     & 14.02  & 56.67             & 9.91    & 46.68              & 38.04    & 87.78              \\
 \bottomrule
\end{tabular}}
\caption{Comparison for BM25 and the dense retrieval model DE w/ BM25 Neg (DE) in the Zero-shot (ZS.) experiments, and fine-tuning (FT.) experiments for estimating the upper-bound performance on cCOVID-News out-of-domain testing set.}
\label{tab:out-of-domain-eval-covid}
\end{table}

\subsection{Out-of-Domain Evaluation} \label{sec:out-of-domain-experiment}
We evaluate the out-of-domain (OOD) generalization ability of our dense retriever (DE w/ BM25 Neg) on the two OOD testing sets. We report the results in two settings: 1) Zero-shot setting: we directly evaluate DE w/ BM25 Neg without fine-tuning. 2) Fine-tuning setting: we fine-tune DE w/ BM25 Neg with the data from the target domain and evaluate it on OOD testing sets. The performance of the fine-tuned models is the estimated upper-bound that DE w/ BM25 Neg can achieve on OOD testing sets.

In Table \ref{tab:out-of-domain-eval-cmedqa} and \ref{tab:out-of-domain-eval-covid}, we summarize the results of the out-of-domain experiments. First, we notice that the performance of the dense retriever is largely degraded on the two OOD testing sets. 
According to the in-domain evaluation (see Table \ref{tab:main-eval-retrieval} and \ref{tab:main-eval-rerank}), the dense retriever considerably outperforms BM25, however it has no obvious advantage over BM25 in the zero-shot setting, or even worse. 
In addition, the dense retriever can be significantly improved by fine-tuning. Its can maintain a large advantage over BM25 after fine-tuning on the target-domain. This results show that the dense retriever has limited domain transfer capability as observed in \cite{thakur2021beir}.

\subsection{Cross-lingual Evaluation} \label{sec:cross-lingual-experiment}

\begin{table}[]
\centering
\scalebox{0.85}{
\begin{tabular}{lcc}
\toprule
\textbf{Model \textbackslash{} Evaluation}                  & \textbf{Monolingual} & \textbf{Cross-lingual}    \\ \midrule
Supervised Model       & -             & 28.03 \\ 
Zero-shot Model & 87.88             & 19.50  \\
Transferred Model & -             & \textbf{38.35} \\ \bottomrule
\end{tabular}}
\caption{\textbf{Monolingual} (retrieving Chinese passages with Chinese queries) and \textbf{Cross-lingual} (retrieving Chinese passages with English queries) performance of the dual-encoder retrievers on our cross-lingual evaluations. We report the Recall@50 score for each retrieval model.}
\label{tab:cross-lingual-eval}
\end{table}

In the cross-lingual evaluation, 
we experiment with three dense retrieval models based on multilingual BERT (mBERT) \cite{devlin-etal-2019-bert}. 
\begin{itemize}
    \item \textbf{Supervised Model} We directly fine-tune mBERT using the parallel data of English queries and Chinese passages. 
    \item \textbf{Zero-shot Model} We fine-tune an mBERT retriever on the full monolingual Chinese training data (i.e., 86K Chinese queries with Chinese positive paragraphs in DuReader$_{\bf retrieval}$), and directly evaluate it on the cross-lingual testing set.
    \item \textbf{Transferred Model} We further fine-tune \textbf{Zero-shot Model} by using the parallel data paired with English queries and Chinese passages, and then evaluate it on the cross-lingual testing set. 
\end{itemize}

As shown in Table \ref{tab:cross-lingual-eval}, we note that the performance of Zero-shot Model on cross-lingual testing set is less effective than Supervised Model. 
Furthermore, Zero-shot Model performs significantly worse on cross-lingual data than on monolingual data. 
According to these findings, cross-lingual retrieval is more difficult than monolingual retrieval, since the retriever cannot find relevant passages by simply matching shared terms between queries and passages \cite{litschko2021cross-lingual-retrieval}. 
Instead, cross-lingual retrievers must capture the semantic relevance of the query and passages. 
Additionally, Transferred Model outperformed other baselines, demonstrating the validity of transferring knowledge from the monolingual Chinese annotated data.

\section{Related Works} 

\noindent \textbf{Passage Retrieval Benchmarks.} Passage retrieval and open-domain question-answering are challenging tasks that attracts much attention in developing the benchmarks. MS-MARCO \cite{nguyen2016msMARCO} contains queries extracted from the search log of Microsoft Bing, which poses challenges in both the retrieval of relevant contexts and reading comprehension based on the contexts. Natural Questions \citep{kwiatkowski-etal-2019-natural-question} is an open-domain question answering benchmark that consist of real queries issued to the Google search engine. These datasets are widely used for the research of passage retrieval. However, \citet{lewis-etal-2021-TestTrainOverlapODQA} find that there are 30\% of test-set queries have semantically overlaps in the training queries for Natural Questions. 
\citet{arabzadeh2021shallowPoolingSparseLabel} observe that false negatives are common in MS-MARCO. 
TianGong-PDR \cite{wu2019investigating} and Sougou-QCL \cite{10.1145/3209978.3210092-sougou-qcl-clickingmodel} are two Chinese retrieval datasets for the news documents and web-pages, separately. However, these datasets are either small or have no human annotation. Despite the progress in developing benchmarks for English passage retrieval, the large-scale and high-quality benchmarks for the non-English community are still limited.

\noindent \textbf{Dense Retrieval Model.} Information retrieval is a long-standing problem. In contrast to the traditional sparse retrieval methods \cite{SALTON1988513-tfidf-term-weighting-text-retrieved,robertson2009probabilisticRelevanceFramework-bm25}, recent dense retrievers aim at encoding the query and retrieved documents as contextualized representations based on the pre-training language models \cite{devlin-etal-2019-bert,sun2019ernie}, then calculate the relevance based on similarity function \cite{karpukhin-etal-2020-dense-passage-retrieval,10.1162/tacl_a_00369-yuan-li-sparse-dense-and-attentional-representation-for-text-retrieval,qu-etal-2021-rocketqa} (e.g. cosine or dot product). Based on different learning paradigms, neural retrieval systems can be divided into two categories: 1) \textit{unsupervised}: pre-training the retrieval without annotated data \cite{chang2019pretraining-largescale-retrieval,DBLP:journals/corr/abs-2108-05540-unsupervised-corpus-model-pre-training}; 2) \textit{supervised}: training the query and document encoders by contrasting the positives with designed negatives  \cite{karpukhin-etal-2020-dense-passage-retrieval,DBLP:conf/iclr/XiongXLTLBAO21-ANCE,10.1145/3404835.3462880-optimizing-dense-retriever-with-hard-negatives}. In terms of system architecture, the recent systems typically follow the two-stage framework (retrieval-then-re-ranking), in which a retriever \cite{mao-etal-2021-retriever-augmented-retrieval-generation,DBLP:journals/corr/abs-1904-08375-document-expansion-retrieval,10.1145/3331184.3331303-deep-text-understand-for-IR-retriever} first retrieve a list of top candidates and the re-ranker \cite{gao-etal-2020-modularized-reranker,10.1145/3397271.3401075-colbert-reranker} will re-rank retrieved candidates.
It has been shown that large-scale annotated datasets are one of the keys to successfully train dense retrievers~\cite{karpukhin-etal-2020-dense-passage-retrieval}.

\section{Conclusion}
This paper presents a large-scale Chinese passage retrieval dataset to benchmark the retrieval systems.
In order to ensure the quality of our dataset, we employ two strategies: 1) reducing the false negatives in development and testing sets using a pooling approach and human annotations, and 2) removing the training queries that are semantically similar to the development and testing queries. In addition, we provide two testing sets for out-of-domain evaluation, and a set for cross-lingual evaluation. We examine several retrieval baselines, including the traditional sparse retrieval system and the neural retrievers, and present the challenges and the limitations of our dataset. We hope this dataset can help facilitate the research of passage retrieval. 

\section{Limitations}

As we discussed in Section \ref{sec:in-domain-challenges}, we still observe that approximately 14.8\% of our best re-ranking model's wrong predictions are indeed caused by false negatives, even though we observed that our quality improvement strategy discussed in Section \ref{sec:quality-improvement} was effective. 
This is primarily due to the difficulty of annotating the training data in a way that captures all positives.

Secondly, two out-of-domain testing sets are restricted to the medical domain. 
cMedQA focuses on the medical question-answering conversations, and cCOVID-NEWS focuses on the medical news domain. 
It may limit the ability to evaluate retrieval systems in other domains (e.g., the financial or legal domains). 

\section{Ethical Consideration}

Our DuReader$_{\bf retrieval}$ is developed only for research purpose. All data is collected from either the open-source projects, respecting corresponding licences' restrictions, or publicly available benchmarks. We do not guarantee that we have the copyright of this data, any may further discard data resources without copyright if necessary.

\section{Acknowledgements}

We thank the anonymous reviewers for their helpful comments. We also appreciate the feedback received from other members of the Deep Question Answering Team at Baidu NLP, specifically Qifei WU, for his valuable discussions and comments. This work is supported by the National Key Research and Development Project of China (No.2018AAA0101900).


\bibliographystyle{acl_natbib}
\bibliography{custom}

\begin{thebibliography}{43}
\expandafter\ifx\csname natexlab\endcsname\relax\def\natexlab#1{#1}\fi

\bibitem[{Arabzadeh et~al.(2021)Arabzadeh, Vtyurina, Yan, and
  Clarke}]{arabzadeh2021shallowPoolingSparseLabel}
Negar Arabzadeh, Alexandra Vtyurina, Xinyi Yan, and Charles~LA Clarke. 2021.
\newblock \href {https://arxiv.org/abs/2109.00062} {Shallow pooling for sparse
  labels}.
\newblock \emph{ArXiv preprint}, abs/2109.00062.

\bibitem[{Asai et~al.(2021{\natexlab{a}})Asai, Kasai, Clark, Lee, Choi, and
  Hajishirzi}]{asai-etal-2021-xor-tidy}
Akari Asai, Jungo Kasai, Jonathan Clark, Kenton Lee, Eunsol Choi, and Hannaneh
  Hajishirzi. 2021{\natexlab{a}}.
\newblock \href {https://doi.org/10.18653/v1/2021.naacl-main.46} {{XOR} {QA}:
  Cross-lingual open-retrieval question answering}.
\newblock In \emph{Proceedings of the 2021 Conference of the North American
  Chapter of the Association for Computational Linguistics: Human Language
  Technologies}, pages 547--564, Online. Association for Computational
  Linguistics.

\bibitem[{Asai et~al.(2021{\natexlab{b}})Asai, Yu, Kasai, and
  Hajishirzi}]{NEURIPS2021_3df07fda-one-question-answering-model-from-many-language-crosslingual}
Akari Asai, Xinyan Yu, Jungo Kasai, and Hanna Hajishirzi. 2021{\natexlab{b}}.
\newblock \href
  {https://proceedings.neurips.cc/paper/2021/file/3df07fdae1ab273a967aaa1d355b8bb6-Paper.pdf}
  {One question answering model for many languages with cross-lingual dense
  passage retrieval}.
\newblock In \emph{Advances in Neural Information Processing Systems},
  volume~34, pages 7547--7560. Curran Associates, Inc.

\bibitem[{Berant et~al.(2013)Berant, Chou, Frostig, and
  Liang}]{berant-etal-2013-webquestions}
Jonathan Berant, Andrew Chou, Roy Frostig, and Percy Liang. 2013.
\newblock \href {https://aclanthology.org/D13-1160} {Semantic parsing on
  {F}reebase from question-answer pairs}.
\newblock In \emph{Proceedings of the 2013 Conference on Empirical Methods in
  Natural Language Processing}, pages 1533--1544, Seattle, Washington, USA.
  Association for Computational Linguistics.

\bibitem[{Bonifacio et~al.(2021)Bonifacio, Jeronymo, Abonizio, Campiotti,
  Fadaee, , Lotufo, and Nogueira}]{bonifacio2021mmarco}
Luiz~Henrique Bonifacio, Vitor Jeronymo, Hugo~Queiroz Abonizio, Israel
  Campiotti, Marzieh Fadaee, , Roberto Lotufo, and Rodrigo Nogueira. 2021.
\newblock \href {http://arxiv.org/abs/2108.13897} {mmarco: A multilingual
  version of ms marco passage ranking dataset}.

\bibitem[{Chang et~al.(2020)Chang, Yu, Chang, Yang, and
  Kumar}]{chang2019pretraining-largescale-retrieval}
Wei{-}Cheng Chang, Felix~X. Yu, Yin{-}Wen Chang, Yiming Yang, and Sanjiv Kumar.
  2020.
\newblock \href {https://openreview.net/forum?id=rkg-mA4FDr} {Pre-training
  tasks for embedding-based large-scale retrieval}.
\newblock In \emph{8th International Conference on Learning Representations,
  {ICLR} 2020, Addis Ababa, Ethiopia, April 26-30, 2020}. OpenReview.net.

\bibitem[{Chen et~al.(2021)Chen, Lakhotia, Oguz, Gupta, Lewis, Peshterliev,
  Mehdad, Gupta, and
  Yih}]{DBLP:journals/corr/abs-2110-06918-salient-phrase-aware-dense-retrieval}
Xilun Chen, Kushal Lakhotia, Barlas Oguz, Anchit Gupta, Patrick S.~H. Lewis,
  Stan Peshterliev, Yashar Mehdad, Sonal Gupta, and Wen{-}tau Yih. 2021.
\newblock \href {https://arxiv.org/abs/2110.06918} {Salient phrase aware dense
  retrieval: Can a dense retriever imitate a sparse one?}
\newblock \emph{ArXiv preprint}, abs/2110.06918.

\bibitem[{Cui et~al.(2020)Cui, Che, Liu, Qin, Wang, and
  Hu}]{cui-etal-2020-revisiting-MacBERT}
Yiming Cui, Wanxiang Che, Ting Liu, Bing Qin, Shijin Wang, and Guoping Hu.
  2020.
\newblock \href {https://doi.org/10.18653/v1/2020.findings-emnlp.58}
  {Revisiting pre-trained models for {C}hinese natural language processing}.
\newblock In \emph{Findings of the Association for Computational Linguistics:
  EMNLP 2020}, pages 657--668, Online. Association for Computational
  Linguistics.

\bibitem[{Cui et~al.(2021)Cui, Che, Liu, Qin, and
  Yang}]{cui-etal-2021-pretrain}
Yiming Cui, Wanxiang Che, Ting Liu, Bing Qin, and Ziqing Yang. 2021.
\newblock \href {https://doi.org/10.1109/TASLP.2021.3124365} {Pre-training with
  whole word masking for chinese bert}.

\bibitem[{Dai and
  Callan(2019)}]{10.1145/3331184.3331303-deep-text-understand-for-IR-retriever}
Zhuyun Dai and Jamie Callan. 2019.
\newblock \href {https://doi.org/10.1145/3331184.3331303} {Deeper text
  understanding for {IR} with contextual neural language modeling}.
\newblock In \emph{Proceedings of the 42nd International {ACM} {SIGIR}
  Conference on Research and Development in Information Retrieval, {SIGIR}
  2019, Paris, France, July 21-25, 2019}, pages 985--988. {ACM}.

\bibitem[{Dang et~al.(2013)Dang, Bendersky, and
  Croft}]{10.1007/978-3-642-36973-5_36-two-stage-learning-to-rank}
Van Dang, Michael Bendersky, and W.~Bruce Croft. 2013.
\newblock \href {https://doi.org/10.1007/978-3-642-36973-5_36} {Two-stage
  learning to rank for information retrieval}.
\newblock In \emph{Proceedings of the 35th European Conference on Advances in
  Information Retrieval}, ECIR'13, pages 423--434, Berlin, Heidelberg.
  Springer-Verlag.

\bibitem[{Devlin et~al.(2019)Devlin, Chang, Lee, and
  Toutanova}]{devlin-etal-2019-bert}
Jacob Devlin, Ming-Wei Chang, Kenton Lee, and Kristina Toutanova. 2019.
\newblock \href {https://doi.org/10.18653/v1/N19-1423} {{BERT}: Pre-training of
  deep bidirectional transformers for language understanding}.
\newblock In \emph{Proceedings of the 2019 Conference of the North {A}merican
  Chapter of the Association for Computational Linguistics: Human Language
  Technologies, Volume 1 (Long and Short Papers)}, pages 4171--4186,
  Minneapolis, Minnesota. Association for Computational Linguistics.

\bibitem[{Gao and
  Callan(2021)}]{DBLP:journals/corr/abs-2108-05540-unsupervised-corpus-model-pre-training}
Luyu Gao and Jamie Callan. 2021.
\newblock \href {https://arxiv.org/abs/2108.05540} {Unsupervised corpus aware
  language model pre-training for dense passage retrieval}.
\newblock \emph{ArXiv preprint}, abs/2108.05540.

\bibitem[{Gao et~al.(2020)Gao, Dai, and
  Callan}]{gao-etal-2020-modularized-reranker}
Luyu Gao, Zhuyun Dai, and Jamie Callan. 2020.
\newblock \href {https://doi.org/10.18653/v1/2020.emnlp-main.342} {Modularized
  transfomer-based ranking framework}.
\newblock In \emph{Proceedings of the 2020 Conference on Empirical Methods in
  Natural Language Processing (EMNLP)}, pages 4180--4190, Online. Association
  for Computational Linguistics.

\bibitem[{He et~al.(2018)He, Liu, Liu, Lyu, Zhao, Xiao, Liu, Wang, Wu, She,
  Liu, Wu, and Wang}]{he-etal-2018-dureader}
Wei He, Kai Liu, Jing Liu, Yajuan Lyu, Shiqi Zhao, Xinyan Xiao, Yuan Liu,
  Yizhong Wang, Hua Wu, Qiaoqiao She, Xuan Liu, Tian Wu, and Haifeng Wang.
  2018.
\newblock \href {https://doi.org/10.18653/v1/W18-2605} {{D}u{R}eader: a
  {C}hinese machine reading comprehension dataset from real-world
  applications}.
\newblock In \emph{Proceedings of the Workshop on Machine Reading for Question
  Answering}, pages 37--46, Melbourne, Australia. Association for Computational
  Linguistics.

\bibitem[{Johnson et~al.(2019)Johnson, Douze, and
  J{\'e}gou}]{johnson2019billion}
Jeff Johnson, Matthijs Douze, and Herv{\'e} J{\'e}gou. 2019.
\newblock Billion-scale similarity search with gpus.
\newblock \emph{IEEE Transactions on Big Data}, 7(3):535--547.

\bibitem[{Joshi et~al.(2017)Joshi, Choi, Weld, and
  Zettlemoyer}]{joshi-etal-2017-triviaqa}
Mandar Joshi, Eunsol Choi, Daniel Weld, and Luke Zettlemoyer. 2017.
\newblock \href {https://doi.org/10.18653/v1/P17-1147} {{T}rivia{QA}: A large
  scale distantly supervised challenge dataset for reading comprehension}.
\newblock In \emph{Proceedings of the 55th Annual Meeting of the Association
  for Computational Linguistics (Volume 1: Long Papers)}, pages 1601--1611,
  Vancouver, Canada. Association for Computational Linguistics.

\bibitem[{Karpukhin et~al.(2020)Karpukhin, Oguz, Min, Lewis, Wu, Edunov, Chen,
  and Yih}]{karpukhin-etal-2020-dense-passage-retrieval}
Vladimir Karpukhin, Barlas Oguz, Sewon Min, Patrick Lewis, Ledell Wu, Sergey
  Edunov, Danqi Chen, and Wen-tau Yih. 2020.
\newblock \href {https://doi.org/10.18653/v1/2020.emnlp-main.550} {Dense
  passage retrieval for open-domain question answering}.
\newblock In \emph{Proceedings of the 2020 Conference on Empirical Methods in
  Natural Language Processing (EMNLP)}, pages 6769--6781, Online. Association
  for Computational Linguistics.

\bibitem[{Khattab and Zaharia(2020)}]{10.1145/3397271.3401075-colbert-reranker}
Omar Khattab and Matei Zaharia. 2020.
\newblock \href {https://doi.org/10.1145/3397271.3401075} {Colbert: Efficient
  and effective passage search via contextualized late interaction over
  {BERT}}.
\newblock In \emph{Proceedings of the 43rd International {ACM} {SIGIR}
  conference on research and development in Information Retrieval, {SIGIR}
  2020, Virtual Event, China, July 25-30, 2020}, pages 39--48. {ACM}.

\bibitem[{Kwiatkowski et~al.(2019)Kwiatkowski, Palomaki, Redfield, Collins,
  Parikh, Alberti, Epstein, Polosukhin, Devlin, Lee, Toutanova, Jones, Kelcey,
  Chang, Dai, Uszkoreit, Le, and
  Petrov}]{kwiatkowski-etal-2019-natural-question}
Tom Kwiatkowski, Jennimaria Palomaki, Olivia Redfield, Michael Collins, Ankur
  Parikh, Chris Alberti, Danielle Epstein, Illia Polosukhin, Jacob Devlin,
  Kenton Lee, Kristina Toutanova, Llion Jones, Matthew Kelcey, Ming-Wei Chang,
  Andrew~M. Dai, Jakob Uszkoreit, Quoc Le, and Slav Petrov. 2019.
\newblock \href {https://doi.org/10.1162/tacl_a_00276} {Natural questions: A
  benchmark for question answering research}.
\newblock \emph{Transactions of the Association for Computational Linguistics},
  7:452--466.

\bibitem[{Lewis et~al.(2021)Lewis, Stenetorp, and
  Riedel}]{lewis-etal-2021-TestTrainOverlapODQA}
Patrick Lewis, Pontus Stenetorp, and Sebastian Riedel. 2021.
\newblock \href {https://doi.org/10.18653/v1/2021.eacl-main.86} {Question and
  answer test-train overlap in open-domain question answering datasets}.
\newblock In \emph{Proceedings of the 16th Conference of the European Chapter
  of the Association for Computational Linguistics: Main Volume}, pages
  1000--1008, Online. Association for Computational Linguistics.

\bibitem[{Litschko et~al.(2021)Litschko, Vuli{\'c}, Ponzetto, and
  Glava{\v{s}}}]{litschko2021cross-lingual-retrieval}
Robert Litschko, Ivan Vuli{\'c}, Simone~Paolo Ponzetto, and Goran Glava{\v{s}}.
  2021.
\newblock \href {https://arxiv.org/abs/2112.11031} {On cross-lingual retrieval
  with multilingual text encoders}.
\newblock \emph{ArXiv preprint}, abs/2112.11031.

\bibitem[{Liu et~al.(2019)Liu, Ott, Goyal, Du, Joshi, Chen, Levy, Lewis,
  Zettlemoyer, and Stoyanov}]{DBLP:journals/corr/abs-1907-11692RoBERTa}
Yinhan Liu, Myle Ott, Naman Goyal, Jingfei Du, Mandar Joshi, Danqi Chen, Omer
  Levy, Mike Lewis, Luke Zettlemoyer, and Veselin Stoyanov. 2019.
\newblock \href {https://arxiv.org/abs/1907.11692} {Roberta: {A} robustly
  optimized {BERT} pretraining approach}.
\newblock \emph{ArXiv preprint}, abs/1907.11692.

\bibitem[{Luan et~al.(2021)Luan, Eisenstein, Toutanova, and
  Collins}]{10.1162/tacl_a_00369-yuan-li-sparse-dense-and-attentional-representation-for-text-retrieval}
Yi~Luan, Jacob Eisenstein, Kristina Toutanova, and Michael Collins. 2021.
\newblock \href {https://doi.org/10.1162/tacl_a_00369} {Sparse, dense, and
  attentional representations for text retrieval}.
\newblock \emph{Transactions of the Association for Computational Linguistics},
  9:329--345.

\bibitem[{Ma et~al.(2019)Ma, Yu, Wu, and Wang}]{ma2019paddlepaddle}
Yanjun Ma, Dianhai Yu, Tian Wu, and Haifeng Wang. 2019.
\newblock Paddlepaddle: An open-source deep learning platform from industrial
  practice.
\newblock \emph{Frontiers of Data and Domputing}, 1(1):105--115.

\bibitem[{Mao et~al.(2021)Mao, He, Liu, Shen, Gao, Han, and
  Chen}]{mao-etal-2021-retriever-augmented-retrieval-generation}
Yuning Mao, Pengcheng He, Xiaodong Liu, Yelong Shen, Jianfeng Gao, Jiawei Han,
  and Weizhu Chen. 2021.
\newblock \href {https://doi.org/10.18653/v1/2021.acl-long.316}
  {Generation-augmented retrieval for open-domain question answering}.
\newblock In \emph{Proceedings of the 59th Annual Meeting of the Association
  for Computational Linguistics and the 11th International Joint Conference on
  Natural Language Processing (Volume 1: Long Papers)}, pages 4089--4100,
  Online. Association for Computational Linguistics.

\bibitem[{Nguyen et~al.(2016)Nguyen, Rosenberg, Song, Gao, Tiwary, Majumder,
  and Deng}]{nguyen2016msMARCO}
Tri Nguyen, Mir Rosenberg, Xia Song, Jianfeng Gao, Saurabh Tiwary, Rangan
  Majumder, and Li~Deng. 2016.
\newblock \href
  {https://www.microsoft.com/en-us/research/publication/ms-marco-human-generated-machine-reading-comprehension-dataset/}
  {Ms marco: A human generated machine reading comprehension dataset}.

\bibitem[{Nogueira et~al.(2019)Nogueira, Yang, Lin, and
  Cho}]{DBLP:journals/corr/abs-1904-08375-document-expansion-retrieval}
Rodrigo Nogueira, Wei Yang, Jimmy Lin, and Kyunghyun Cho. 2019.
\newblock \href {https://arxiv.org/abs/1904.08375} {Document expansion by query
  prediction}.
\newblock \emph{ArXiv preprint}, abs/1904.08375.

\bibitem[{Qu et~al.(2021)Qu, Ding, Liu, Liu, Ren, Zhao, Dong, Wu, and
  Wang}]{qu-etal-2021-rocketqa}
Yingqi Qu, Yuchen Ding, Jing Liu, Kai Liu, Ruiyang Ren, Wayne~Xin Zhao, Daxiang
  Dong, Hua Wu, and Haifeng Wang. 2021.
\newblock \href {https://doi.org/10.18653/v1/2021.naacl-main.466}
  {{R}ocket{QA}: An optimized training approach to dense passage retrieval for
  open-domain question answering}.
\newblock In \emph{Proceedings of the 2021 Conference of the North American
  Chapter of the Association for Computational Linguistics: Human Language
  Technologies}, pages 5835--5847, Online. Association for Computational
  Linguistics.

\bibitem[{Robertson and
  Zaragoza(2009)}]{robertson2009probabilisticRelevanceFramework-bm25}
Stephen Robertson and Hugo Zaragoza. 2009.
\newblock The probabilistic relevance framework: Bm25 and beyond.
\newblock \emph{Information Retrieval}, 3(4):333--389.

\bibitem[{Salton and
  Buckley(1988)}]{SALTON1988513-tfidf-term-weighting-text-retrieved}
Gerard Salton and Christopher Buckley. 1988.
\newblock \href {https://doi.org/https://doi.org/10.1016/0306-4573(88)90021-0}
  {Term-weighting approaches in automatic text retrieval}.
\newblock \emph{Information Processing \& Management}, 24(5):513--523.

\bibitem[{Shi et~al.(2021)Shi, Zhang, Bai, and
  Lin}]{shi-etal-2021-crosslingual-dense-document-retrieval}
Peng Shi, Rui Zhang, He~Bai, and Jimmy Lin. 2021.
\newblock \href {https://doi.org/10.18653/v1/2021.mrl-1.24} {Cross-lingual
  training of dense retrievers for document retrieval}.
\newblock In \emph{Proceedings of the 1st Workshop on Multilingual
  Representation Learning}, pages 251--253, Punta Cana, Dominican Republic.
  Association for Computational Linguistics.

\bibitem[{Sun and Duh(2020)}]{sun-duh-2020-clirmatrix}
Shuo Sun and Kevin Duh. 2020.
\newblock \href {https://doi.org/10.18653/v1/2020.emnlp-main.340}
  {{CLIRM}atrix: A massively large collection of bilingual and multilingual
  datasets for cross-lingual information retrieval}.
\newblock In \emph{Proceedings of the 2020 Conference on Empirical Methods in
  Natural Language Processing (EMNLP)}, pages 4160--4170, Online. Association
  for Computational Linguistics.

\bibitem[{Sun et~al.(2019)Sun, Wang, Li, Feng, Chen, Zhang, Tian, Zhu, Tian,
  and Wu}]{sun2019ernie}
Yu~Sun, Shuohuan Wang, Yukun Li, Shikun Feng, Xuyi Chen, Han Zhang, Xin Tian,
  Danxiang Zhu, Hao Tian, and Hua Wu. 2019.
\newblock \href {https://arxiv.org/abs/1904.09223} {Ernie: Enhanced
  representation through knowledge integration}.
\newblock \emph{ArXiv preprint}, abs/1904.09223.

\bibitem[{Thakur et~al.(2021)Thakur, Reimers, R{\"u}ckl{\'e}, Srivastava, and
  Gurevych}]{thakur2021beir}
Nandan Thakur, Nils Reimers, Andreas R{\"u}ckl{\'e}, Abhishek Srivastava, and
  Iryna Gurevych. 2021.
\newblock Beir: A heterogeneous benchmark for zero-shot evaluation of
  information retrieval models.
\newblock In \emph{Thirty-fifth Conference on Neural Information Processing
  Systems Datasets and Benchmarks Track (Round 2)}.

\bibitem[{Voorhees et~al.(2005)Voorhees, Harman et~al.}]{voorhees2005trec}
Ellen~M Voorhees, Donna~K Harman, et~al. 2005.
\newblock \emph{TREC: Experiment and evaluation in information retrieval},
  volume~63.
\newblock Citeseer.

\bibitem[{Wu et~al.(2019)Wu, Mao, Liu, Zhang, and Ma}]{wu2019investigating}
Zhijing Wu, Jiaxin Mao, Yiqun Liu, Min Zhang, and Shaoping Ma. 2019.
\newblock \href {https://doi.org/10.1145/3331184.3331233} {Investigating
  passage-level relevance and its role in document-level relevance judgment}.
\newblock In \emph{Proceedings of the 42nd International {ACM} {SIGIR}
  Conference on Research and Development in Information Retrieval, {SIGIR}
  2019, Paris, France, July 21-25, 2019}, pages 605--614. {ACM}.

\bibitem[{Xiong et~al.(2021)Xiong, Xiong, Li, Tang, Liu, Bennett, Ahmed, and
  Overwijk}]{DBLP:conf/iclr/XiongXLTLBAO21-ANCE}
Lee Xiong, Chenyan Xiong, Ye~Li, Kwok{-}Fung Tang, Jialin Liu, Paul~N. Bennett,
  Junaid Ahmed, and Arnold Overwijk. 2021.
\newblock \href {https://openreview.net/forum?id=zeFrfgyZln} {Approximate
  nearest neighbor negative contrastive learning for dense text retrieval}.
\newblock In \emph{9th International Conference on Learning Representations,
  {ICLR} 2021, Virtual Event, Austria, May 3-7, 2021}. OpenReview.net.

\bibitem[{Zhan et~al.(2021)Zhan, Mao, Liu, Guo, Zhang, and
  Ma}]{10.1145/3404835.3462880-optimizing-dense-retriever-with-hard-negatives}
Jingtao Zhan, Jiaxin Mao, Yiqun Liu, Jiafeng Guo, Min Zhang, and Shaoping Ma.
  2021.
\newblock \href {https://doi.org/10.1145/3404835.3462880} {\emph{Optimizing
  Dense Retrieval Model Training with Hard Negatives}}, pages 1503--1512.
  Association for Computing Machinery, New York, NY, USA.

\bibitem[{Zhan et~al.(2022)Zhan, Xie, Mao, Liu, Zhang, and
  Ma}]{Zhan2022EvaluatingEP}
Jingtao Zhan, Xiaohui Xie, Jiaxin Mao, Yiqun Liu, M.~Zhang, and Shaoping Ma.
  2022.
\newblock Evaluating extrapolation performance of dense retrieval.
\newblock \emph{ArXiv}, abs/2204.11447.

\bibitem[{Zhang et~al.(2018)Zhang, Zhang, Wang, Guo, and Liu}]{cmedq}
Sheng Zhang, Xin Zhang, Hui Wang, Lixiang Guo, and Shanshan Liu. 2018.
\newblock \href {https://doi.org/10.1109/ACCESS.2018.2883637} {Multi-scale
  attentive interaction networks for chinese medical question answer
  selection}.
\newblock \emph{IEEE Access}, 6:74061--74071.

\bibitem[{Zheng et~al.(2018)Zheng, Fan, Liu, Luo, Zhang, and
  Ma}]{10.1145/3209978.3210092-sougou-qcl-clickingmodel}
Yukun Zheng, Zhen Fan, Yiqun Liu, Cheng Luo, Min Zhang, and Shaoping Ma. 2018.
\newblock \href {https://doi.org/10.1145/3209978.3210092} {Sogou-qcl: {A} new
  dataset with click relevance label}.
\newblock In \emph{The 41st International {ACM} {SIGIR} Conference on Research
  {\&} Development in Information Retrieval, {SIGIR} 2018, Ann Arbor, MI, USA,
  July 08-12, 2018}, pages 1117--1120. {ACM}.

\bibitem[{Zhu et~al.(2021)Zhu, Chen, Yan, Liu, Hong, Chen, Wu, and
  Wang}]{zhu2021duqm}
Hongyu Zhu, Yan Chen, Jing Yan, Jing Liu, Yu~Hong, Ying Chen, Hua Wu, and
  Haifeng Wang. 2021.
\newblock \href {https://arxiv.org/abs/2112.08609} {Duqm: A chinese dataset of
  linguistically perturbed natural questions for evaluating the robustness of
  question matching models}.
\newblock \emph{ArXiv preprint}, abs/2112.08609.

\end{thebibliography}

\newpage

\appendix

\section{Appendix}
\label{sec:appendix}

\subsection{Details for Re-ranker Used in Reducing False Negatives}
\label{appendix:pooling-details}
We first use four different pre-training models, including ERNIE \cite{sun2019ernie}, BERT \cite{devlin-etal-2019-bert, cui-etal-2021-pretrain}, RoBERTa \cite{DBLP:journals/corr/abs-1907-11692RoBERTa} and MacBERT \cite{cui-etal-2020-revisiting-MacBERT}, as the initializations to train four cross-encoder re-rankers as in \cite{qu-etal-2021-rocketqa} with negatives sampled from the pooled passages as discussed in Section \ref{sec:quality-improvement}. We then ensemble these four re-ranking models by averaging their prediction scores for each query-passage pair. 

\subsection{Baseline Implementation Details}
\label{appendix:implmentation-details}
We conduct all experiments with the deep learning framework PaddlePaddle \cite{ma2019paddlepaddle} on up to eight NVIDIA Tesla A100 GPUs (with 40G RAM).

We use the ERNIE 1.0 base \cite{sun2019ernie} as the initializations for both our first dual-encoder retriever (DE w/ BM25 Neg) and cross-encoder re-ranker (CE w/ BM25 Neg). ERNIE shares the same architecture with BERT but is trained with entity-level masking and phrase-level masking to obtain better knowledge-enhanced representations. To train our second enhanced re-ranker (CE w/ DE Neg), we use the parameters from CE w/ BM25 Neg as initialization. 

For training settings, we also use the Cross-batch negatives setting as in \cite{qu-etal-2021-rocketqa}. When sampling the hard negatives from the top-50 retrieved items, we sample 4 negatives per positive passage. The dual-encoders are trained with the batch size of 256. The cross-encoders
are trained with the batch size of 64. The dual-encoders and cross-encoders are trained with 10 and 3 epochs. We use ADAM optimizer for all models' trainings and the learning rate of the dual-encoder is set to 3e-5 with the rate of linear scheduling warm-up at 0.1, while the learning rate of the cross-encoder is set to 1e-5 with no warm-up training. We set the maximal length of questions and passages as 32 and 384, respectively.

In inference time of our dense retrieval model (DE w/ BM25 Neg), we use FAISS \cite{johnson2019billion} to index the dense representations of all passages.

\subsection{Details for Human Annotations}
\label{appendix:crosslingual-set-details}
We perform the annotation in our internal annotation platform to ensure the data quality, where all the annotators and reviewers are full-time employees. The pairs of all queries and their pooled top-5 paragraphs retrieved by all models are divided into packages, with 1K samples for each. Annotators are asked to identify whether each query-paragraph pair is relevant for a single package. Then at least two reviewers check the accuracy of this package by reviewing 100 random query-paragraph pairs independently. If the average accuracy is less than the threshold (i.e., 93\%), the annotators will be asked to revise the package until the accuracy is higher than the threshold.

\subsection{Cases for Challenges in Error Analysis}\label{sec:case-error-analysis}

We present the selected cases in Table \ref{tab:error-analysis} and discuss them in this section to support our error analysis in Section \ref{sec:in-domain-challenges}.

\noindent \textbf{Salient Phrase Mismatch} \quad Taking the entity mismatch as an example, we expect that the main entity in the retrieved passage should be consistent with the query. However, the second example in Table \ref{tab:error-analysis} shows that the query asks for information about \textit{Taobao}, but the retrieved passage is related to \textit{Alipay} instead. There is a challenge for retrieval systems to filter out passages that entail entities inconsistent with the query.

\noindent \textbf{Syntax Mismatch} \quad Given the case showed in Table \ref{tab:error-analysis} as an example, the retrieval system is hard to understand the subject and object in the example query are \textit{Taipei} and \textit{Ruifang}, instead, it simply ranks the candidate passage entailing \textit{Taipei} and \textit{Ruifang} to a top predictions.

\noindent \textbf{Other Challenges} \quad In our analysis, it is found that about 21\% of the errors are due to the 
retrieval system simply predicting its output based on the presence of co-occurring low-frequency terms (e.g., "\textit{wow}" in the example in Table \ref{tab:error-analysis}) in query and paragraph, but their semantic meanings are not related indeed. And about 1.6\% of the errors are due to noise in the query or paragraph. For example, misspelling the "\textit{iPhone}" as "\textit{ipone}".

\begin{table*}[]
\centering
\scalebox{0.65}{
\begin{tabular}{c p{2cm} p{3cm} p{6cm} p{6cm} c}
\toprule \toprule
\textbf{Category}       & \textbf{Type}           & \textbf{Example Query}      & \textbf{Example Passage }                                       & \textbf{Explanation}      & \textbf{$\%$}\\ \midrule
\multirow{3}*{Salient Phrase Mismatch} &    Entity mismatch                & \begin{CJK} {UTF8}{gbsn}{\color[HTML]{9a0018}{淘宝}}修改实名认证\end{CJK} 

Change authentication name at {\color[HTML]{9a0018}{Taobao}} 
& \begin{CJK}{UTF8}{gbsn}响应国家规定，{\color[HTML]{9a0018}{支付宝}}即将实施支付宝个人信息实名认证...\end{CJK}

In response to national regulations, {\color[HTML]{9a0018}{Alipay}} will soon implement real-name authentication...
&       The entities in the query (Taobao) and the passage (Alipay) are mismatched.            &     39\%       \\ \cmidrule{2-6}
~   & Numeral mismatch    & \begin{CJK}{UTF8}{gbsn}最近有什么好听的歌{\color[HTML]{9a0018}{2016}}\end{CJK}       

Any nice songs in {\color[HTML]{9a0018}{2016}}
& \begin{CJK}{UTF8}{gbsn}...音乐巴士{\color[HTML]{9a0018}{2017}}好听的歌榜单收藏了你最需要的的好歌...\end{CJK}                      

...Music Bus's Best Songs {\color[HTML]{9a0018}{2017}} has the songs you need most...
& The query asks for songs in 2016, but the passage is about songs in 2017.             &       5\%     \\ \cmidrule{2-6}
~   & Modifier mismatch & \begin{CJK}{UTF8}{gbsn}{\color[HTML]{9a0018}{吃完海鲜}}可以喝牛奶吗\end{CJK}

Can I drink milk {\color[HTML]{9a0018}{after having seafood}}?
& \begin{CJK}{UTF8}{gbsn}不可以。{\color[HTML]{9a0018}{早晨}}喝牛奶，不科学。原因是…\end{CJK} 

No. Drinking milk {\color[HTML]{9a0018}{in the morning}} is not good for health. The reason is...
&      The modifier in query (after eating seafood) and the one in the passage (in the morning) are different.            &     9.4\%       \\ \midrule
Syntactic mismatch & Syntactic mismatch        & \begin{CJK}{UTF8}{gbsn}{\color[HTML]{1633b5}{台北}}怎么去{\color[HTML]{1633b5}{瑞芳}}\end{CJK}

How to go from {\color[HTML]{1633b5}{Taipei}} to {\color[HTML]{1633b5}{Ruifang}}
& \begin{CJK}{UTF8}{gbsn}从{\color[HTML]{1633b5}{瑞芳}}回{\color[HTML]{1633b5}{台北}}，乘火车1小时到达，可以使用悠游卡...\end{CJK}  

It takes 1 hour by train from {\color[HTML]{1633b5}{Ruifang}} to {\color[HTML]{1633b5}{Taipei}}. You can use the EasyCard...
&     The query asks how to go from Taipei to Ruifang but the paragraph is about going from Ruifang to Taipei.               &     1\%       \\ \midrule
\multirow{2}*{Other Challenges} & Robustness on typos                & \begin{CJK}{UTF8}{gbsn}{\color[HTML]{26731c}{\sout{ipone}}iPhone手机}屏幕右上角有个圈是什么\end{CJK} 

What is the circle in the upper right corner of the {\color[HTML]{26731c}{\sout{ipone}iPhone}} screen
& \begin{CJK}{UTF8}{gbsn}但是又不知道这些图标是怎么出现的。是什么东西，干什么用的，例如手机信号格旁边的眼睛图标是代表了开启只能屏幕…\end{CJK} 

But I don't know how these icons appeared. What is it and what is it for? For example, the eye icon next to the mobile phone signal grid means that the screen can only be turned on...
&       Typos may introduce noise to the model's understanding of query or passage, e.g., it may affects the identification of the main entity iPhone in the query.           &    1.6\%        \\ 

\cmidrule{2-6}

~ 
& Over-sensitivity on term overlap          
& \begin{CJK}{UTF8}{gbsn}{\color[HTML]{26731c}{wow邮件}}发错了怎么办\end{CJK}

What should I do if my wow email is sent by mistake
& \begin{CJK}{UTF8}{gbsn}...还有个方法就是你登陆{\color[HTML]{26731c}{WOW}}然后都设置好后退出游戏…\end{CJK} 

...and another way is that you log in to {\color[HTML]{26731c}{WOW}} and then exit the game after setting everything up...
&         The query and passage (i.e., WOW) has a matched term in ; but they are semantically irrelevant.         &     21\%   \\ \midrule 

\multirow{2}*{Others} & False negatives                & \begin{CJK}{UTF8}{gbsn}求推荐好看的国产电视剧\end{CJK} 

Please recommend good Chinese TV series
& \begin{CJK}{UTF8}{gbsn}TOP3：美人制造。30集全。主演：杨蓉金世佳邓萃雯。简介：以唐代女皇武则天时期为大背景...\end{CJK} 

TOP3: Beauty Made. 30 episodes. Starring: Yang Rong, Jin Shijia, Deng Cuiwen. Introduction: Taking the period of Empress Wu Zetian of the Tang Dynasty as the background...
                                         & \textbackslash{} &      14.8\%      \\ \cmidrule{2-6} 
~ & Others                           & \textbackslash{}   & \textbackslash{}                                       & Cases not fitting into any of above categories. &      8.2\%     \\
\bottomrule \bottomrule
11R\end{tabular}}
\caption{Summary of the manual analysis for the 500 query-passage pairs predicted by our strongest re-ranker (CE w/ DE Neg). We highlight the challenges in \textit{salient phrase mismatch} in {\color[HTML]{9a0018}{red}}, \textit{syntax mismatch} in {\color[HTML]{1633b5}{blue}}, and \textit{Other Challenges} in {\color[HTML]{26731c}{green}}.}
\label{tab:error-analysis}
\end{table*}

\end{document}